\documentclass[runningheads]{llncs}
\usepackage[T1]{fontenc}
\usepackage{amsmath}
\usepackage{booktabs}
\usepackage{graphicx}
\usepackage[colorlinks=true, allcolors=blue]{hyperref}
\begin{document}

\title{Mask-Guided Multi-Channel SwinUNETR Framework for Robust MRI Classification}

\titlerunning{SwinUNETR Framework for Robust MRI Classification}
\author{Smriti Joshi\inst{1} \and
Lidia Garrucho \inst{1}\and
Richard Osuala \inst{1} \and
Oliver Diaz \inst{1, 2} \and
Karim Lekadir\inst{1, 3}}

\authorrunning{S. Joshi et al.}

\institute{Departament de Matemàtiques i Informàtica, Universitat de Barcelona, Spain
\email{smriti.joshi@ub.edu}
\and
Computer Vision Center, Bellaterra, Spain
\and
Institució Catalana de Recerca i Estudis Avançats (ICREA), Passeig Lluís Companys 23, Barcelona, Spain
}
\maketitle            

\begin{abstract}
Breast cancer is one of the leading causes of cancer-related mortality in women, and early detection is essential for improving outcomes. Magnetic resonance imaging (MRI) is a highly sensitive tool for breast cancer detection, particularly in women at high risk or with dense breast tissue, where mammography is less effective. The ODELIA consortium organized a multi-center challenge to foster AI-based solutions for breast cancer diagnosis and classification. The dataset included 511 studies from six European centers, acquired on scanners from multiple vendors at both 1.5 T and 3 T. Each study was labeled for the left and right breast as no lesion, benign lesion, or malignant lesion. We developed a SwinUNETR-based deep learning framework that incorporates breast region masking, extensive data augmentation, and ensemble learning to improve robustness and generalizability. Our method achieved second place on the challenge leaderboard, highlighting its potential to support clinical breast MRI interpretation. We publicly share our codebase at \url{https://github.com/smriti-joshi/bcnaim-odelia-challenge.git}.

\end{abstract}
\section{Introduction}
 To-date, breast cancer is the most common cancer worldwide, affecting one in eight women in their lifetime \cite{globalCancerObservatory}. Early detection is crucial for improving patient outcomes. While mammography is widely used for screening, its sensitivity is limited in women with dense breast tissue or high-risk profiles. Magnetic resonance imaging (MRI) has emerged as a highly sensitive modality for detecting breast lesions, providing both anatomical and functional information through dynamic contrast-enhanced (DCE) sequences. However, the increasing use of breast MRI generates large volumes of data, placing a significant burden on radiologists and highlighting the need for automated interpretation tools. 
% % \cite{merchant1993advantages}\cite{francca2017role}\cite{metcalfe2013potential}. 

Deep learning methods have shown great promise in medical image analysis. By leveraging spatiotemporal patterns of contrast uptake, AI models can potentially differentiate benign from malignant lesions \cite{macura2006patterns}, improve diagnostic accuracy, and assist radiologists in clinical decision-making. Nevertheless, the development of robust AI models for breast MRI faces several challenges, including inter-center variability, heterogeneous acquisition protocols \cite{joshi2025singleimagetesttimeadaptation}, and class imbalance.

The ODELIA consortium, in collaboration with MICCAI 2025 and the Deep-Breath workshop, organized a multi-center ODELIA challenge\footnote{https://odelia2025.grand-challenge.org/} to advance AI-based breast cancer classification from MRI. The task required participants to develop generalizable algorithms capable of detecting malignancies in breast MRIs by classifying each breast as normal, benign, or malignant. In this work, we present a framework that integrates breast mask–guided preprocessing, multi-channel dynamic contrast–enhanced (DCE) MRI inputs, and a transformer-based SwinUNETR backbone with ensemble learning to address this challenge. The proposed approach leverages both spatial and temporal patterns in DCE-MRI, incorporates strategies to mitigate class imbalance, and delivers robust predictions across heterogeneous multi-center datasets. Our results highlight the potential of AI-driven methods to support radiologists in early and accurate breast cancer detection.

\section{Method}

\subsection{Input Construction}
\begin{figure}
    \centering
    \includegraphics[width=\textwidth]{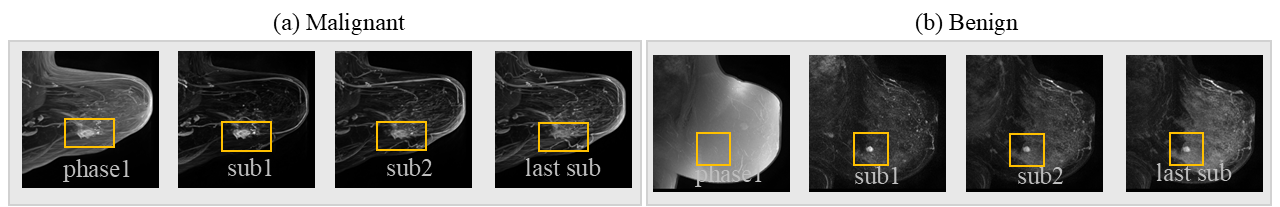}
    \caption{Representative examples of maximum intensity projections (MIPs) showing contrast enhancement kinetics in breast MRI. Phase 1 corresponds to the first post-contrast phase of the DCE-MRI. Subtraction images (sub\textbf{x}) denote the $x^{th}$ post-contrast phase minus the pre-contrast phase. The last phase is dynamically selected for each case, as the number of post-contrast phases varies. The left panel demonstrates a malignant lesion, while the right panel demonstrates a benign lesion. These images illustrate the temporal patterns of contrast uptake that assist in differentiating benign from malignant lesions.}
    \label{fig:contrast}
    
\end{figure}

The input to the network consists of four-channel maximum intensity projections (MIPs) derived from dynamic contrast-enhanced MRI: the first post-contrast phase, subtraction 1 (first post-contrast minus pre-contrast), subtraction 2 (second post-contrast minus pre-contrast), and the last subtraction image. Prior to feeding these channels into the network, each image was masked using the predicted breast segmentation, ensuring that only breast tissue contributed to the feature extraction process. This multi-channel input captures both the baseline enhancement and the temporal dynamics of contrast uptake, which are key discriminative features in breast MRI. Malignant lesions typically show rapid initial enhancement and delayed washout, while benign lesions often display slower, progressive enhancement \cite{macura2006patterns}. This can also be seen in the representative example in Figure \ref{fig:contrast}. Including the first post-contrast image alongside the subtraction images ensures that cases where the tumor is already visible in the pre-contrast or fat-suppressed image are not disadvantaged. By providing the network with multiple temporal contrasts and their subtractions from baseline, the model can more effectively learn patterns of enhancement kinetics, improving its ability to differentiate between benign and malignant lesions.

\subsection{Architecture}
We employed a SwinUNETR model \cite{hatamizadeh2021swin} from MONAI \cite{cardoso2022monai}\footnote{\url{https://docs.monai.io/en/1.3.0/_modules/monai/networks/nets/swin_unetr.html}}
 as the backbone architecture for breast MRI segmentation. SwinUNETR combines the advantages of transformer-based global attention with a U-Net\cite{ronneberger2015u} style encoder-decoder, enabling the model to capture both local texture and long-range dependencies MRI data. The classification head was implemented as a lightweight module attached to the encoder output of the SwinUNETR backbone. It consisted of an adaptive average pooling layer that aggregated spatial features into a single representation, followed by a flattening operation to convert the pooled features into a vector. Finally, a fully connected linear layer mapped the feature vector to the output classes: normal, benign, malignant.

To handle the inherent class imbalance in the dataset, the model is trained with two strategies: one model with natural class weighting and one with additional re-weighting to emphasize underrepresented classes. These are computed through inverse frequency computation for each class. 
Let $N_c$ denote the number of samples in class $c$, and $C$ be the total number of classes. 
The weight $w_c$ for class $c$ is calculated as:

\begin{equation}
w_c = \frac{\frac{1}{N_c}}{\sum_{i=1}^{C} \frac{1}{N_i}}
\end{equation}

These weights were applied to the Cross Entropy Loss during training, giving higher importance
to underrepresented classes while ensuring normalized contributions across all classes as follows:

\begin{equation}
\mathcal{L}_{\text{WCE}} = - \frac{1}{N} \sum_{i=1}^{N} \sum_{c=1}^{C} w_c \, y_{i,c} \, \log(\hat{y}_{i,c})
\end{equation}

where $y_{i,c}$ is the one-hot encoded ground-truth label for sample $i$ and  $\hat{y}_{i,c}$ is the predicted probability for sample $i$. Finally, the models trained with and without weights in cross entropy loss are ensembled to improve robustness and reduce variance.

\section{Experiments and Results}

\subsection{Dataset}

\begin{figure}
    \centering
    \includegraphics[width=\textwidth]{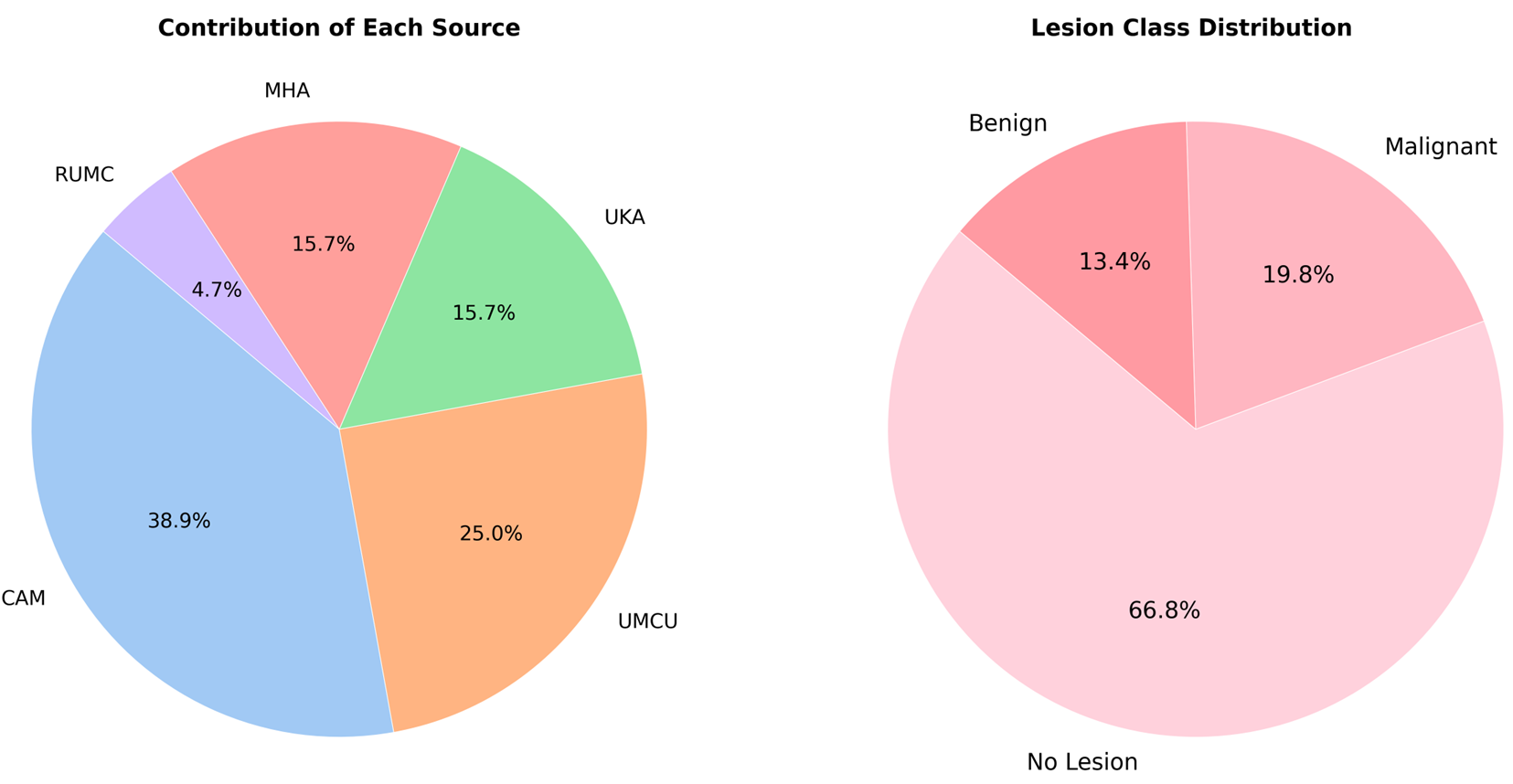}
    \caption{Overview of the ODELIA challenge dataset. Left: Contribution of different centers to the total dataset, illustrating the multi-center and heterogeneous nature of the data.  Right: Class distribution across all breast labels, showing the relative proportions of “no lesion,” “benign lesion,” and “malignant lesion.”
    \\ \textit{Abbreviations}: \textbf{CAM}: Cambridge University Hospitals, Cambridge, UK; \textbf{MHA}: Mitera Hospital, Athens, Greece; \textbf{RSH}: Ribera Hospital, Valencia, Spain; \textbf{RUMC}: Radboud University Medical Center, Nijmegen, Netherlands; \textbf{UKA}: University Hospital Aachen, Aachen, Germany; \textbf{UMCU}: University Medical Center Utrecht, Utrecht, Netherlands.}
    \label{fig:dataset}
\end{figure}
Two distinct datasets were used in this work: one for training a breast mask segmentation model and one for developing the breast cancer classification model.

\textbf{Segmentation:} To generate accurate breast masks, we trained a segmentation model using publicly available databases. Specifically, we used:

\begin{enumerate}
    \item \textbf{Duke-Breast-Cancer-MRI} \cite{duke_dataset}: This dataset was originally collected at Duke University and contains breast DCE-MRI studies from patients with biopsy-confirmed breast cancer. It includes multiple imaging sequences such as pre-contrast and post-contrast T1-weighted images, along with associated clinical information. Importantly, a subset of this dataset provides manually delineated breast masks, which can be used for whole-breast segmentation tasks. For our study, we used 36 cases from this subset where breast masks were available.
    \item \textbf{Breast-Cancer-DCE-MRI} \cite{zhang2023breast_cancer_dce_mri_data}: This dataset contains a cohort of 100 cases from Yunnan Cancer Hospital, each comprising breast DCE-MRIs with six phases (one pre-contrast and five post-contrast images) with corresponding whole-breast segmentations.
\end{enumerate}

\textbf{Classification:} The breast cancer classification model was trained solely on the dataset provided by the ODELIA challenge \cite{müllerfranzes2025europeanmulticenterbreastcancer}. This dataset contains 511 breast MRI studies, each corresponding to a unique patient. Every study was assigned two labels, one for the left breast and one for the right breast, with possible categories being no lesion, benign lesion, or malignant lesion. The dataset is heterogeneous, having been curated across six European centers using scanners from multiple vendors at both 1.5 T and 3 T field strengths. Each study included a T2-weighted acquisition alongside a DCE-MRI acquisition consisting of one pre-contrast phase and between two and seven post-contrast phases. 

An overview of the dataset composition is shown in Figure \ref{fig:dataset}. The contribution of individual centers reflects the multi-center and heterogeneous nature of the challenge data, spanning different vendors and acquisition protocols. As shown on the right, the class distribution is imbalanced, with a noticeably higher proportion of no lesion cases compared to benign and malignant categories. 

\subsection{Preprocessing Pipeline}
All breast MRI studies underwent a standardized preprocessing pipeline to ensure spatial consistency and extract informative inputs for classification. For each case, four temporal phases were selected: the pre-contrast image, the first and second post-contrast images, and the last available post-contrast image. Each volume was loaded using TorchIO \cite{P_rez_Garc_a_2021}, reoriented to canonical space, and resampled to a fixed voxel spacing of 0.7 × 0.7 × 3 $mm^3$. The images were then cropped or padded to a uniform size of 512 × 512 × 32 voxels and subsequently, the height was cropped to 256 through an intensity-based localization used to retain the breast region. The volumes were then divided into left and right breast halves at 50\% width. Finally, the maximum intensity projections (MIPs) are computed along the z-axis and normalized.

\subsection{Implementation Details}
The segmentation model was trained for 100 epochs with an initial 5-epoch warm-up phase, using a batch size of 20 and a learning rate of $0.0001$, scheduled with cosine annealing. Using the same optimization strategy, the classification model was trained for 300 epochs with a batch size of 10. Extensive data augmentation was applied in both networks using Albumentations \cite{buslaev2020albumentations}, including random horizontal and vertical flips, rotations, affine transformations with scaling, shearing and translation, as well as elastic deformations, grid and optical distortions, and random resized crops. Intensity augmentations included random brightness and contrast adjustments, multiplicative noise, Gaussian blur, motion blur, Gaussian noise, and coarse dropout to simulate missing signal. All inputs were normalized using channel-wise means (0.2074, 0.1290, 0.1396, 0.1470) and standard deviations (0.2110, 0.1629, 0.1620, 0.1626).

We performed five-fold cross-validation stratified by patient-level lesion labels. First, the maximum lesion label for each patient was computed, e.g. if the left breast is benign and right breast is malignant, the patient is assigned malignant label, only for stratification purposes. Using these labels, patients were partitioned into five folds with StratifiedKFold from scikit-learn \cite{pedregosa2011scikit}. For each fold, the training and validation sets were defined at the patient level, preventing data leakage between folds.

\begin{table}[h]
\centering
\caption{\textbf{Quantitative result.} Reported metrics include area under the receiver operating characteristic curve (AUC), sensitivity at 90\% specificity, and specificity at 90\% sensitivity. The overall score is defined as the average of these three metrics and was used to rank the leaderboard. Results are shown for both cross-validation and held-out test sets from the challenge.}
\begin{tabular}{lccccc}
\toprule
Test Set & Weights & AUC    & Sensitivity & Specificity & Score \\    
\midrule   
Fold 1    & No      & 0.8670 & 0.6707      & 0.5915     & 0.7097   \\
Fold 1    & Yes     & 0.8072 & 0.4939      & 0.4054     & 0.5688    \\
Fold 2    & No      & 0.9078 & 0.7427      & 0.7256     & 0.7920   \\
Fold 2    & Yes     & 0.8580 & 0.5060      & 0.6280     & 0.6640   \\
Fold 3    & No      & 0.8769 & 0.6890      & 0.6311     & 0.7323   \\
Fold 3    & Yes     & 0.7578 & 0.3720      & 0.3567     & 0.4955   \\
Fold 4    & No      & 0.8887 & 0.7593      & 0.5864     & 0.7448   \\
Fold 4    & Yes     & 0.8551 & 0.7222      & 0.4599     & 0.6791   \\
Fold 5    & No      & 0.8797 & 0.7099      & 0.64814    & 0.7459    \\
Fold 5    & Yes     & 0.8116 & 0.5309      & 0.4383     & 0.5936 \\
\midrule
Test set  &Ensem. & 0.8610 & 0.6201      & 0.5678     & 0.6830 \\
\bottomrule
\end{tabular}
\label{tab:results}
\end{table}

\begin{figure}
    \centering
    \includegraphics[width=\textwidth]{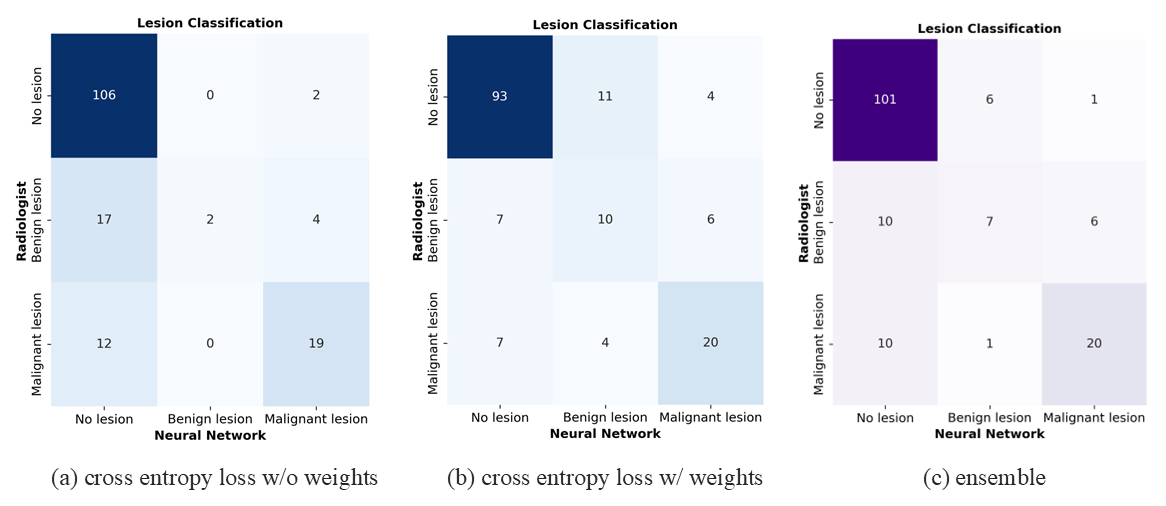}
    \caption{Confusion matrices comparing classification performance with natural class weighting, inverse-frequency class weighting, and their ensemble. Class weighting improves benign detection but increases false positives for normal cases. Ensembles strikes a balance between the two approaches.}
    \label{fig:confusion_matrix}
\end{figure}
\subsection{Results}

\textbf{Cross-Validation Performance:} We evaluated the classification framework using five-fold cross-validation, comparing models trained with natural class weighting versus inverse-frequency class weighting. Table \ref{tab:results} summarizes the micro-AUC, sensitivity at 90\% specificity, and specificity at 90\% sensitivity for both strategies. Incorporating class weights generally improved performance on underrepresented classes, particularly benign lesions, but increased false predictions in the majority normal class. Consequently, the overall micro-AUC slightly decreased, reflecting the dominance of normal cases in the dataset. Models trained without weighting achieved higher micro-AUC by favoring the majority class, at the cost of reduced sensitivity for benign and malignant cases—a trend observed consistently across all folds. Figure \ref{fig:confusion_matrix} shows this behavior through the confusion matrices in fold-4. To balance these tradeoffs, predictions from both strategies were combined through ensembling the predicted probabilities. The ensemble improved correct classification of malignant lesions, though a tradeoff remained between benign and normal cases due to their similar imaging characteristics, underscoring the challenge of distinguishing subtle benign lesions from normal tissue in dynamic contrast-enhanced MRI. The final model, obtained by ensembling across all folds, achieved an AUC of 0.86 on the held-out challenge test set.

\begin{figure} 
\centering
    \includegraphics[width=0.6\textwidth]{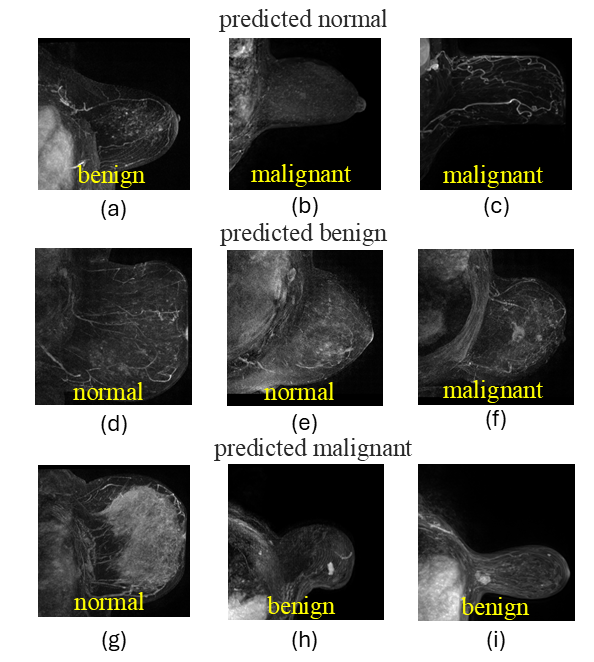}
    \caption{Examples of misclassified cases. The ground-truth label is overlaid on each corresponding image.}
    \label{fig:failure_cases}
\end{figure}

\textbf{Qualitative Analysis of Failure Cases:} Representative failure cases are shown in Figure \ref{fig:failure_cases}. Misclassifications occurred primarily between normal and benign cases due to a similar presentations (see Figure \ref{fig:failure_cases}a. and Figure \ref{fig:failure_cases}e.) , low signal-to-noise ratio (Figure \ref{fig:failure_cases}b.), or small lesions (see Figure \ref{fig:failure_cases}c.). In some cases, benign lesions were predicted as malignant and vice versa (see Figure \ref{fig:failure_cases}f. and see Figure \ref{fig:failure_cases}i), illustrating the inherent difficulty of differentiating these classes. 

\section{Discussion}

In this study, we developed and evaluated a multi-channel deep learning framework for breast lesion classification from dynamic contrast-enhanced (DCE) MRI. By using maximum intensity projections of the first post-contrast phase together with sequential subtraction images, the network may have been able to capture both morphological and kinetic information relevant for case differentiation. This could have contributed to the observed discriminability between benign and malignant tumors. The integration of a segmentation-guided classification strategy provided a tissue-focused representation that excluded background signal, reducing noise and improving discriminative power.

These findings also highlight the fundamental difficulty in distinguishing benign from normal breast tissue, which often exhibit similar post-contrast dynamics. While malignant lesions typically demonstrate stronger and more rapid enhancement, benign findings can overlap substantially, leading to misclassification. Future research may benefit from integrating additional modalities such as diffusion-weighted imaging \cite{dimitriadis2025assessing} or radiomics-derived features \cite{lambin2017radiomics}, which may provide complementary information beyond contrast kinetics.

\begin{credits}
\subsubsection{\ackname}
This project has received funding from European research and innovation programme under grant agreement No 101057699 (RadioVal) and Horizon 2020 under grant agreement No 952103 (EuCanImage). The work has also been supported by FUTURE-ES (PID2021-126724OB-I00) and AIMED (PID2023-146786OB-I00) from the Ministry of Science and Innovation of Spain.

\subsubsection{\discintname}
%\subsubsection*{Disclosure of Interests.} 
The authors have no competing interests to declare that are relevant to the content of this article.
\end{credits}

%
% ---- Bibliography ----
\bibliography{references} % bibliography data in report.bib
\bibliographystyle{splncs04}

\end{document}